\documentclass[lettersize,journal]{IEEEtran}
\usepackage{amsmath,amsfonts}
\usepackage{algorithmic}
\usepackage{algorithm}
\usepackage{array}
\usepackage[caption=false,font=normalsize,labelfont=sf,textfont=sf]{subfig}
\usepackage{textcomp}
\usepackage{stfloats}
\usepackage{url}
\usepackage{verbatim}
\usepackage{graphicx}
\usepackage{subfloat}
\usepackage{multirow}
\usepackage{cite}
\usepackage{amssymb}
\usepackage[section]{placeins}

\makeatletter

\newcommand{\Rmnum}[1]{\expandafter\@slowromancap\romannumeral #1@}
\makeatother

\begin{document}

\title{SAR-NeRF: Neural Radiance Fields for Synthetic Aperture Radar Multi-View Representation}

\author{Zhengxin Lei,~\IEEEmembership{Graduate Student Member,~IEEE},
        Feng Xu,~\IEEEmembership{Senior Member,~IEEE},
        Jiangtao Wei,~\IEEEmembership{Graduate Student Member,~IEEE},
        Feng Cai,~\IEEEmembership{Graduate Student Member,~IEEE},
        Feng Wang,~\IEEEmembership{Member,~IEEE},
        and Ya-Qiu Jin,~\IEEEmembership{Life Fellow,~IEEE}
\thanks{This work was supported by the Natural Science Foundation of China under Grant U2130202. (Corresponding author: Feng Xu.).}
\thanks{The authors are with the Key Laboratory for Information Science of Electromagnetic Waves (MoE), Fudan University, Shanghai 200433, China.(e-mail: fengxu@fudan.edu.cn)}}

\markboth{Journal of \LaTeX\ Class Files,~Vol.~14, No.~8, August~2021}%
{Shell \MakeLowercase{\textit{et al.}}: A Sample Article Using IEEEtran.cls for IEEE Journals}


\maketitle

\begin{abstract}
SAR images are highly sensitive to observation configurations, and they exhibit significant variations across different viewing angles, making it challenging to represent and learn their anisotropic features. As a result, deep learning methods often generalize poorly across different view angles. Inspired by the concept of neural radiance fields (NeRF), this study combines SAR imaging mechanisms with neural networks to propose a novel NeRF model for SAR image generation. Following the mapping and projection pinciples, a set of SAR images is modeled implicitly as a function of attenuation coefficients and scattering intensities in the 3D imaging space through a differentiable rendering equation. SAR-NeRF is then constructed to learn the distribution of attenuation coefficients and scattering intensities of voxels, where the vectorized form of 3D voxel SAR rendering equation and the sampling relationship between the 3D space voxels and the 2D view ray grids are analytically derived. Through quantitative experiments on various datasets, we thoroughly assess the multi-view representation and generalization capabilities of SAR-NeRF. Additionally, it is found that SAR-NeRF augumented dataset can significantly improve SAR target classification performance under few-shot learning setup, where a 10-type classification accuracy of 91.6\% can be achieved by using only 12 images per class.
\end{abstract}


\section{Introduction}
\IEEEPARstart{S}{ynthetic} Aperture Radar (SAR) has been widely utilized in the field of earth remote sensing due to its capability for all-weather and all-day observation. However, the complexity of SAR imagery interpretation poses challenges for deep learning-based methods employed in SAR image classification and target recognition. The performance of these methods is often limited by the diversity and scale of training samples. Additionally, SAR imagery exhibits high sensitivity to observation configurations, resulting in substantial variations between images acquired under different conditions. Particularly, SAR images undergo significant changes with varying viewing angles, making it challenging to characterize and learn their multi-view features effectively. In particular, deep learning approaches are significantly impacted by variations in viewing angles, leading to weaker generalization capabilities across different viewing angles. This further highlights the issue of limited sample availability for SAR-based deep-learning interpretation methods.

To address the challenges of few-shot learning and cross-view generalization in SAR imagery, there are currently two main approaches: transfer learning and new-view generation methods. The main idea behind transfer learning is to pre-train the network using other readily available data that share similar semantic features with SAR imagery and then transfer it to the SAR dataset. Common types of pretraining data include data from other sensors and simulated data. However, the differences between data from different modalities or sources can pose challenges for feature transfer in the network. For instance, pretraining and transferring optical data to SAR imagery may introduce errors due to the fundamental inconsistencies between optical and SAR features\cite{kang2016sar,toizumi2018automatic}. 

Another approach is the new-view sample generation method, which utilizes generative models to train multi-view representation models from existing perspective images. These models can generate new SAR images, thereby augmenting the training dataset with samples from different angles. The first type of method involves training the network through physics-based simulation, known as the EM-simulation method \cite{song2019simulation}, which is often limited by the variety and realism of the simulated data. The other type is pure generative models including Generative Adversarial Networks (GAN) or AutoEncoders, such as adversarial autoencoder\cite{song2021learning}. However, these methods are still primarily data-driven and do not fully incorporate the physical principles of SAR imaging into the network. As a result, they can only learn the interpolation capability between adjacent angles, achieving a smooth transition effect as the viewing angle changes. The effectiveness of these methods in addressing the above-mentioned challenges is limited.

In the field of optical imagery, the NeRF (Neural Radiance Fields)\cite{mildenhall2021nerf} model introduced a method based on implicit representation. By incorporating a physical model of rendering optical 3D voxels into a neural network, the NeRF model achieved dense reconstruction of 3D voxels from multi-view observed images. Furthermore, by utilizing imaging model projection, the NeRF model can successfully generate images of new perspectives, effectively addressing the problem of multi-view image generation in the optical image domain. The NeRF model has demonstrated impressive results in this regard.

This paper presents a new SAR-NeRF model, which is based on the fundamental scattering and imaging mechanism of SAR. It constructs a neural network model for SAR image voxel rendering using mapping and projection algorithms (MPA)\cite{xu2006imaging}. The imaging space is divided into voxels which are then sampled by ray grids from different view angles, enabling the learning of multi-view SAR image representations. Extensive experiments are conducted using rendered synthetic data and measured MSTAR data, accompanied by quantitative evaluations. In summary, the main contributions of this paper can be summarised as follows:

\begin{enumerate}
\item{Constructing SAR neural radiance fields: The SAR neural radiance field method is developed using voxel rendering techniques and the viewpoint-sampling point transformation equation. This method enables the learning of the distribution of attenuation coefficients and scattering intensities in the sampling space through the utilization of multi-view SAR observation data.}
\item{Introducing SAR image voxel rendering and viewpoint-sampling point transformation equation: The paper proposes a SAR image voxel rendering method that is easily integrated with neural networks. Generating voxel distributions based on observed viewpoints effectively partitions the imaging space into voxels.}
\item{We achieved multi-view representation and generation of SAR images and reconstructs target geometry models based on multi-view SAR images. Extensive demonstration and evaluation are conducted, involving a wide range of datasets and numerous experiments. The validation of multi-view SAR image generation is accomplished, showcasing the capability of extracting three-dimensional models based on rendered data.}
\end{enumerate}

The remaining sections of this paper are organized as follows: In Section \Rmnum{2}, a brief introduction is provided on SAR image simulation methods, generative adversarial networks, and neural radiance fields of related works. In Section \Rmnum{3}, a neural network model for SAR image voxel rendering is constructed based on the mapping and projection principles of SAR imaging. The voxel partitioning method is also presented based on observed viewpoints in the imaging space. In Section \Rmnum{4}, experiments are conducted to generate multi-view images using various datasets. Quantitative evaluation metrics are designed, and the extraction of geometric models from multi-view SAR observation images is achieved. Section \Rmnum{5} summarizes the paper, and concluding remarks are provided.

\section{Related works}
\subsection{Physics-Based SAR image simulation methods}
Physics-based SAR image simulation methods are commonly used to simulate real-world environments and can address the challenges encountered in actual scenarios, thereby mitigating the issue of limited SAR image observation samples. These methods can be broadly classified into two categories: coherent echo simulation methods and incoherent image generation methods.

Regarding coherent echo simulation methods, Xu et al. proposed the Bidirectional Ray Tracing (BART) technique for computing the Radar Cross Section (RCS) of large-sized three-dimensional targets with rough surfaces. This method efficiently calculates the RCS of complex scattering scenarios involving large-scale 3D ships on rough sea surfaces, enabling numerical calculations of RCS for both monostatic and Bistatic configurations\cite{xu2009bidirectional}. Yue et al. presented an improved Generalized Gaussian Correlation (GGCS) coherence model to generate coherent SAR images. They introduced adjustments to the scatterer number restrictions and flexible selection of Gaussian scattering distribution parameters, providing a more general and realistic approach for SAR image representation\cite{yue2021coherent}. Zhang et al. proposed a method based on the Fast Beamforming Algorithm (FBAM) and the Gaussian Optics Physical Optics (GO-PO) technique. This method computes the complex scattering of complex ship targets on rough sea surfaces and compares the results with real ship targets, validating the effectiveness of the approach\cite{zhang2016reliable}.

For non-coherent image generation methods, Xu et al. proposed the Mapping Projection method for simulating polarimetric SAR imaging of complex terrain scenes. The expression for SAR imaging of polarimetric scattering in complex scenes is derived and successfully simulated SAR imaging under various configurations\cite{xu2006imaging}. Fu et al. introduced a differentiable renderer that enables forward rendering from 3D models to 2D images and inverse reconstruction from 2D images to 3D models. They demonstrated the feasibility of inverse imaging methods for SAR image generation\cite{fu2022differentiable}. Balz et al. developed a real-time SAR simulation system based on GPU processing. They utilized a rasterization approach for real-time single-bounce simulation, significantly improving the speed of SAR image simulation\cite{balz2009hybrid}.

Note that there is a substantial amount of work in this field, while only a few examples are introduced.

\subsection{Generating Adversarial Networks}
In the domain of SAR image generation, generative adversarial networks (GANs) have been extensively employed to address the issue of generating new azimuth angles in SAR images. Ding et al. proposed a pose generation method that utilizes azimuth interpolation to generate linearly synthesized SAR images with specific azimuth angles\cite{ding2016convolutional}. Subsequently, many studies have utilized the generative capabilities of GANs for data augmentation in SAR images\cite{cui2019image,xie2021data,zheng2019semi,lu2019data}. Liu et al. used CycleGAN to complete the angle of SAR aircraft targets\cite{liu2018sar}. At the same time, Zhang et al. incorporated an azimuth discrimination model into an improved DCGAN to linearly synthesize SAR images with different azimuth angles\cite{zhang2018data}. Oh et al. proposed PeaceGAN to estimate the attitude angles and target class information of SAR target images\cite{oh2021peacegan}]. Although these methods can improve the classification accuracy in target recognition through synthetic samples, significant discrepancies exist between the generated and true images. Additionally, some works have employed other deep neural networks to simulate SAR images. Guo et al. utilized a deep feature transformation method based on differential vectors to generate realistic samples considering labels, azimuth angles, and target characteristics\cite{guo2021deep}. Song et al. introduced the AAE for image generation network, which significantly enhances the recognition accuracy under limited sample conditions\cite{song2021learning}. Dong et al. employed an improved recurrent neural network to model sequence azimuth target images for predicting missing azimuth angle SAR images\cite{dong2022new}. However, the methods above only approach SAR azimuth angle generation from the perspective of image representation using neural networks without considering the actual scattering mechanisms and the image projection geometry of SAR systems. Tushar et al. proposed a pose synthesis method based on sparse modeling of available images in the training data, utilizing the anisotropic scattering behavior of the scattering center of interest related to the viewing angle to simulate nearby attitudes\cite{agarwal2020sparse}. Nevertheless, this method cannot generate SAR images under different pitch angles and requires significant manual annotation costs.

\subsection{Neural Radiance Fields}
The integration of deep learning with relevant data priors to solve related problems has recently sparked a research trend in Implicit Neural Representation (INR) \cite{mescheder2019occupancy}. Neural Radiance Fields (NeRF) applies INR to the task of novel view synthesis in optical images and represents a data-driven approach centered around neural volume rendering\cite{mildenhall2021nerf}. The training process of NeRF involves two main steps: scene encoding and rendering. During the scene encoding phase, NeRF learns the positions and colors of each point in the scene using a set of input images and corresponding camera parameters. It represents each point as a latent vector and employs neural networks to map the input images and camera parameters to these vectors. In the rendering phase, NeRF utilizes the trained model to generate images from new viewpoints. It samples points along each ray and calculates the color and density of each point using the scene encoding network, ultimately producing the final image. A key advantage of NeRF is its ability to generate realistic synthesized images, including rich geometric details and lighting effects. By incorporating the physical principles of the sensor observation into neural networks, NeRF achieves the presentation and re-editing of 3D content from naturally observed scenes, providing a new direction for few-shot methods.

\section{SAR Neural Radiance Fields}
Based on the principles of SAR imaging and mapping projection\cite{xu2006imaging}, this paper first constructs a forward-generation model called SAR Neural Radiance Fields (SAR-NeRF). The model converts the viewpoint information of SAR images, such as radar altitude, azimuth, and pitch angle, into 3D voxel sampling point information. This sampling point information is then input into an $MLP$ encoder to estimate the attenuation coefficient and scattering intensity of the corresponding voxels. Subsequently, the MPA-SAR voxel rendering equation is employed to generate the final SAR image. The flowchart of SAR Neural Radiance Fields is illustrated in Fig. 1. In this section, we will discuss four aspects: the mapping and projection principles of SAR imaging, the construction of the sampling space mapping relationship, SAR image voxel rendering, and learning of the radiance field.
\begin{figure}[!t]
\centering
\includegraphics[width=3.5in]{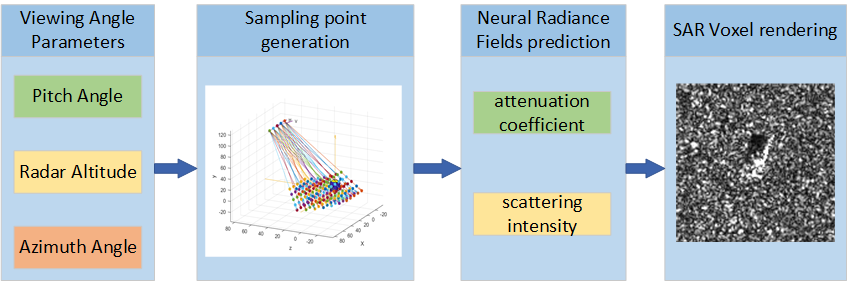}
\caption{SAR neural radiance field flow chart}
\label{fig_1}
\end{figure}

\subsection{SAR 3D voxel rendering equation based on MPA}
SAR achieves two-dimensional high-resolution imaging through pulse compression and synthetic aperture techniques, as illustrated in Fig.2. The platform flies at an altitude $H$ along the $x$-axis direction, continuously emitting signals toward the ground and receiving the scattered echoes from ground targets. Throughout this process, the radar antenna maintains a fixed viewpoint (typically a broadside view). Let $O$ denote the center point of the illuminated area, and $R$ represent the slant range between point $O$ and the radar flight path. $\theta_r$ and $\theta_a$ denote the vertical and azimuth beamwidths, respectively, while $W$ represents the swath of the imaging area, and $L_s$ represents the effective synthetic aperture length.
\begin{figure}[!t]
\centering
\includegraphics[width=3.5in]{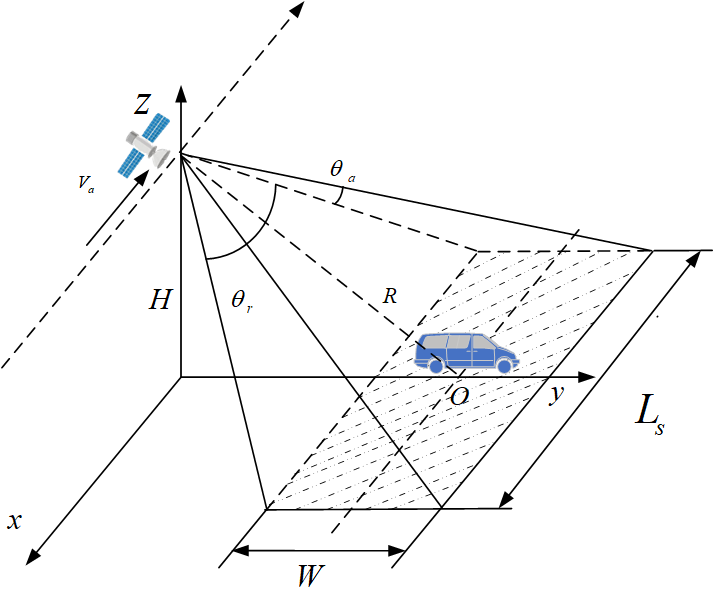}
\caption{SAR imaging geometric model}
\label{fig_2}
\end{figure}

From Fig. 2, it can be observed that the illuminated area of the SAR image is determined by the actual aperture and radiation pattern of the radar antenna. As the radar moves along the azimuth direction, it continuously receives echoes from the targets and scenes and performs imaging through signal processing. The geometric relationship of the strip map imaging in broadside view is determined by parameters such as the orbit altitude, incidence angle, and azimuth angle. Since the radar is far from the targets, the incidence angle $\theta$ variation for the same target can be neglected as the radar moves along the azimuth direction. Therefore, we assume that the scattering contribution from the same target collected at different radar positions is the same. Hence, in the imaging simulation, we calculate the contribution of each row separately within each azimuth resolution interval. The schematic diagram of the mapping and projection is shown in Fig. 3.
\begin{figure}[!t]
\centering
\includegraphics[width=3.5in]{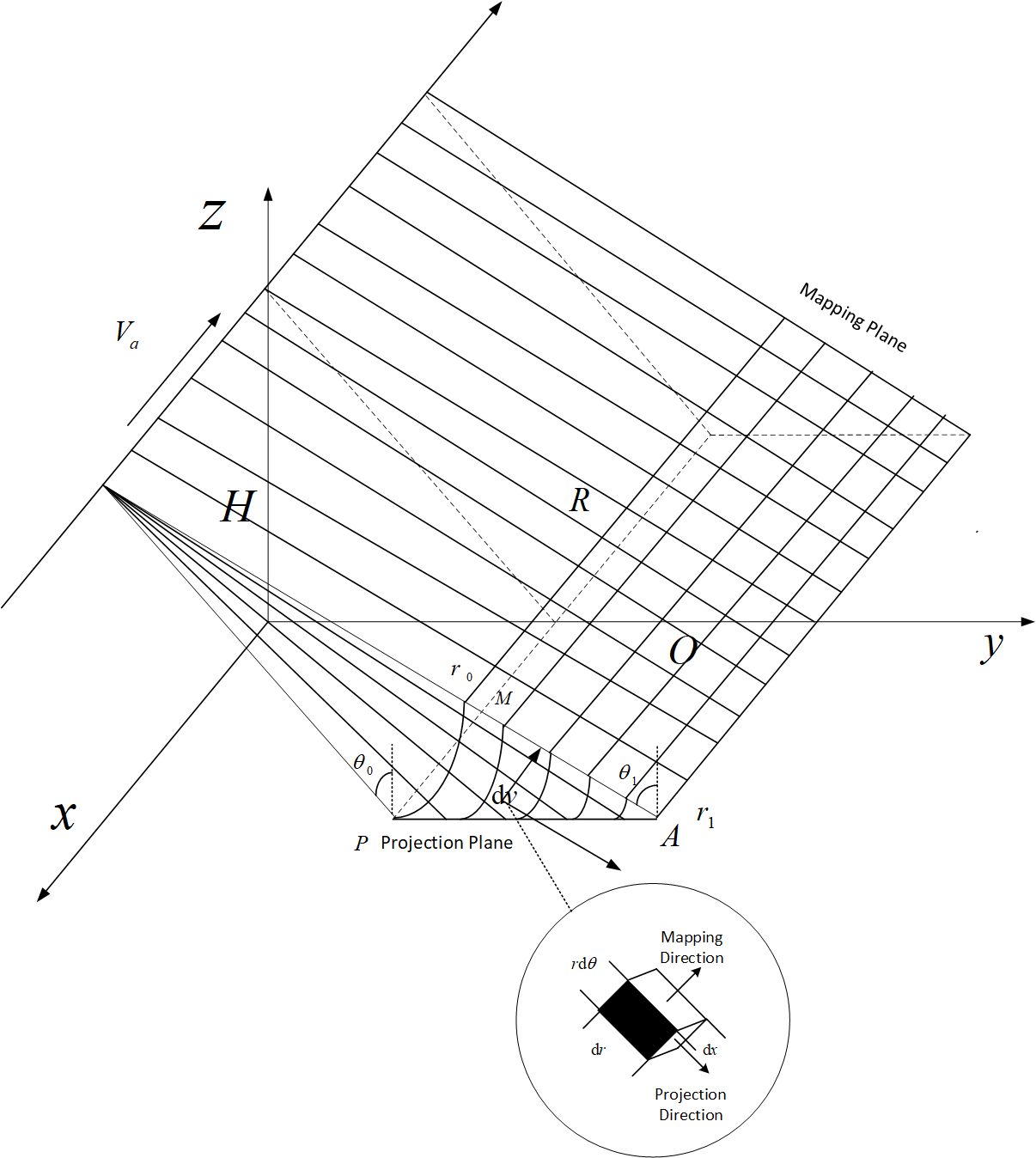}
\caption{Mapping and projection Schema}
\label{fig_3}
\end{figure}

Considering one single cross-section in the azimuth direction as the incident plane, we establish a polar coordinate system $(r,\theta)$ with the radar position as the origin, where $\theta$ represents the radar's incidence angle and $r$ represents the slant range. By determining the sampling range of the radar's received pulse echoes and the range of incidence angle variation, we can define the imaging space of the radar as $r\in\left[r_0,r_1\right]$ and, $\theta\in\left[\theta_0,\theta_1\right]$. Now, let's assume that a single grid unit in the imaging space $\left(x,r,\theta\right)$ corresponds to a voxel ${\rm d} v$, with dimensions ${\rm d}x$,${\rm d}r$,$r{\rm d}\theta$, respectively. According to the radiation transfer theory, when the incident wave $I_i$ passes through a single voxel, the backscattered intensity per unit area, denoted as $I_s$can be expressed as follows\cite{xu2006imaging}:
\begin{equation}
I_s\left(x,r,\theta\right)=E^+\left(x,r,\theta\right)P\left(x,r,\theta\right)E^-\left(x,r,\theta\right)I_i{\rm d}r
\end{equation}
\begin{equation}
E^+\left(x,r,\theta\right)=\exp{\left[-\int_{r_0}^{r}{{\rm d}r^\prime k_e^+\left(x,r^\prime,\theta\right)}\right]}
\end{equation}
\begin{equation}
E^-\left(x,r,\theta\right)=\exp{\left[-\int_{r}^{r_0}{{\rm d}r^\prime k_e^-\left(x,r^\prime,\theta\right)}\right]}
\end{equation}

$E^+$ and $E^-$ represent the accumulated attenuation coefficients in the forward and backward directions. The phase function P denotes the scattering coefficient of the voxel, while $k_e^+\left(x,r,\theta\right)$ and $k_e^-\left(x,r,\theta\right)$ represent the extinction coefficients in the forward and backward directions, respectively. The product of the scattering intensity of the scattering element and the effective penetrating area gives the contribution of scattering energy, as shown in the following equation:
\begin{equation}
S\left(x,r\right){\rm d}x=\int_{\theta_0}^{\theta_1}{I_s\left(x,r,\theta\right)r{\rm d}x{\rm d}\theta}
\end{equation}

By substituting Eqs. (1), (2), and (3) into Eqs. (4) we can obtain the scattering energy of a single pixel in the SAR image, as shown in the following equation:
\begin{equation}
\begin{aligned} 
S_{i,j}&=\iint S\left(x,r\right){\rm d}x{\rm d}r\\
&=\int_{x_i}^{x_{i+1}}\int_{r_j}^{r_{j+1}}\int_{\theta_0}^{\theta_1}
\exp{\left[-\int_{r_0}^{r}{dr^\prime k_e^+\left(x,r^\prime,\theta\right)}\right]}\cdot \\
&\quad P\left(x,r,\theta\right)
\exp{\left[-\int_{r}^{r_0}{dr^\prime k_e^-\left(x,r^\prime,\theta\right)}\right]}r{\rm d}\theta {\rm d}r{\rm d}x
\end{aligned} 
\end{equation}

In natural environments, the random distribution of objects can be quite intricate, making it challenging to derive analytical solutions for the phase function and extinction coefficient in Eqs. (5). Therefore, it is necessary to discretize Eqs. (5) to facilitate its computational treatment. Given the great distance between the radar and the targets, we can use the variable $s$ as a substitute for the pitch angle, where ${\rm d}s=r{\rm d}\theta$. Consequently, we can partition the imaging space $\left(x,r,s\right)$ into a grid by defining $x_m=m\Delta x$,$r_p=p\Delta r$,$s_q=q\Delta s$. This discretization process yields the discrete form of Eqs. (5), as expressed in Eqs. (6). Each grid point in the coordinate system corresponds to a voxel within the imaging space.

\begin{equation}
\begin{aligned} 
S_{i,j}=\Delta x\Delta r\Delta s\sum_{q^{'}=0}^{N_\theta - 1}\prod_{p^{'}=p_0}^{p_j}\exp\left [  -\Delta rk_{e}^{+}\left ( m_i,p^{'},q \right ) \right ]\cdot\\
\quad P\left ( m_i,p_j,q^{'} \right)\prod_{p^{'}=p_j}^{p_0}\exp\left [  -\Delta rk_{e}^{-}\left ( m_i,p^{'},q \right ) \right ] 
\end{aligned}
\end{equation}

The variables $i$ and $j$ correspond to the grid indices of the pixels in the SAR image, while $N_\theta$ represents the number of samples in the scanning angle, with Eqs. (6), we established the model of a three-dimensional voxel rendering method based on the principles of mapping and projection.

\subsection{SAR image 3D voxel rendering equation vectorization}
In the previous section, we derived the three-dimensional voxel rendering equation based on MPA\cite{xu2006imaging}. However, this equation is established in the grid coordinate system, where each voxel has variations in shape and size, making it challenging to integrate with neural networks. Therefore, in this section, we further optimize the voxel rendering equation (equation (6)) by using the center coordinates of the voxels to represent them. Let's assume that the SAR image has dimensions of $N_a\cdot N_r$, and the number of samples in the scanning angle is $N_\theta$. Using the voxel partitioning method from the previous section, we divide the sampling space into $N_a\cdot N_r\cdot N_\theta$ voxels, where the center coordinates of the $\left(i,j,k\right)$-th voxel are denoted as $\left(m_i,p_j,q_k\right)$, the attenuation coefficient at that point is represented as $\sigma_{i,j,k}$, and the scattered intensity along the projection direction is denoted as $S_{i,j,k}$. Based on this, we can obtain the simplified SAR three-dimensional voxel rendering equation, where the scattered intensity of the $\left(i,j\right)$-th pixel unit can be expressed as
\begin{equation}
I_{i,j}=\sum_{k=1}^{N_\theta}{\left(\prod_{j^\prime=1}^{j}e^{jk\sigma_{i,j^\prime,k}}\right)S_{i,j,k}}\left(\prod_{j^\prime=j}^{1}e^{jk\sigma_{i,j^\prime,k}}\right)
\end{equation}

Here for simpling, we choose to ignore the polarimetric characteristics (i.e., assuming forward loss is equivalent to backward loss), Eqs. (7) can be written as follows:
\begin{equation}
I_{i,j}=\sum_{k=1}^{N_\theta}{S_{i,j,k}\left(\prod_{j^\prime=1}^{j}e^{\sigma_{i,j^\prime,k}}\right)}
\end{equation}

For brevity, this paper represents the cumulative multiplication operation in Eqs. (8) using matrix operation. Let's assume $k=k^\prime$, and $I\left(k^\prime\right)$ represents the matrix collection of $I_{i,j}$ when $k=k^\prime$. We can write $I\left(k^\prime\right)$ in matrix form as follows:

\small{
\begin{equation}
\begin{aligned}
\setlength{\arraycolsep}{0.8pt}
&I\left ( k^{'}  \right ) =\\
&\begin{bmatrix} \begin{smallmatrix}
 S_{1,1,k^{'} }\prod_{j^{'}=1\ }^{1}\exp{\left(\sigma_{1,j^{'},k^{'}}\right)}  & \cdots  & S_{1,N_r,k^{'} }\prod_{j^{'}=1\ }^{N_r}\exp{\left(\sigma_{1,j^{'},k^{'}}\right)}\\
\vdots  & \ddots & \vdots\\
 S_{N_a,1,k^{'} }\prod_{j^{'}=1\ }^{1}\exp{\left(\sigma_{N_a,j^{'},k^{'}}\right)} & \cdots  & S_{N_a,N_r,k^{'} }\prod_{j^{'}=1\ }^{N_r}\exp{\left(\sigma_{N_a,j^{'},k^{'}}\right)}
\end{smallmatrix}\end{bmatrix} 
\end{aligned}
\end{equation}}

In this case, we can separate $I\left(k^\prime\right)$ into the product of the scattering intensity matrix and the extinction coefficient matrix: $I\left(k^\prime\right)=S(k^\prime)E(k^\prime)$, where $S(k^\prime)$ and $E(k^\prime)$ are defined as follows:

\begin{equation}
S\left(k^\prime\right)=\left[\begin{matrix}
S_{1,1,k^\prime}&\cdots&S_{1,j,k^\prime}&\cdots&S_{1,N_r,k^\prime}\\
\cdots&\ &\cdots&\ &\cdots\\
S_{i,1,k^\prime}&\cdots&S_{i,j,k^\prime}&\cdots&S_{i,N_r,k^\prime}\\
\cdots&\ &\cdots&\ &\cdots\\
S_{N_a,1,k^\prime}&\cdots&S_{N_a,j,k^\prime}&\cdots&S_{N_a,N_r,k^\prime}\\\end{matrix}\right]
\end{equation}

\begin{equation}
E\left(k^\prime\right)=\mathbf{exp}{\left[ \sigma\left(k^\prime\right)\cdot {\rm TRI}\left(k^\prime\right)\right]}
\end{equation}
\noindent The function $\mathbf{exp}\left(\bullet \right)$ is used to perform an exponential operation on every element in the matrix,  ${\rm TRI}\left(k^\prime\right)$ represents an upper triangular matrix with the same size as $\sigma\left(k^\prime\right)$, and $\sigma(k^\prime)$ are defined as follows:
\begin{equation}
        \sigma\left(k^\prime\right)=\left[\begin{matrix}\sigma_{1,1,k^\prime}&\cdots&\sigma_{1,j,k^\prime}&\cdots&\sigma_{1,N_r,k^\prime}\\\cdots&\ &\cdots&\ &\cdots\\\sigma_{i,1,k^\prime}&\cdots&\sigma_{i,j,k^\prime}&\cdots&\sigma_{i,N_r,k^\prime}\\\cdots&\ &\cdots&\ &\cdots\\\sigma_{N_a,1,k^\prime}&\cdots&\sigma_{N_a,j,k^\prime}&\cdots&\sigma_{N_a,N_r,k^\prime}\\\end{matrix}\right]
\end{equation}

\subsection{Sampling of 3D voxels by 2D ray array}
In the previous section, we derived the three-dimensional voxel rendering equation that depends on the voxel's center coordinates and the projection direction. Therefore, we can calculate the SAR image using the rendering equation by obtaining the center coordinates of all voxels and the projection direction in the imaging space. In this section, we will utilize SAR imaging parameters such as azimuth angle, orbit height, and pitch angle to design a method of sampling the three-dimensional voxels using two-dimensional rays.

In the SAR neural radiance fields, considering that the relative positions between the radar and the targets are different in each SAR image, it is necessary to place the three-dimensional voxels at different observation angles in the same coordinate system. In this study, we define the center of the imaging target as the origin $O$ in the world coordinate system $O-XYZ$. The radar's position, denoted as $O^\prime$, is known and its coordinates are given as  $\overrightarrow{P}_r =\left(x_r,y_r,z_r\right)$. The radar's motion direction is $O^\prime H$, the slant range direction is $O^\prime K$, and the slant off-nadir direction is $O^\prime V$. Thus, a local coordinate system for radar observation, denoted as $O^\prime-KHV$, can be established, where $\overrightarrow{k}$, $\overrightarrow{h}$, and $\overrightarrow{v}$ represent the unit vectors along the $O^\prime K$, $O^\prime H$, and $O^\prime V$ axes, respectively. Fig.4 illustrates the definition of the coordinate system, where $\theta$ and $\varphi$ represent the incident angle and azimuth angle, respectively.

\begin{figure}[!t]
\centering
\includegraphics[width=3.5in]{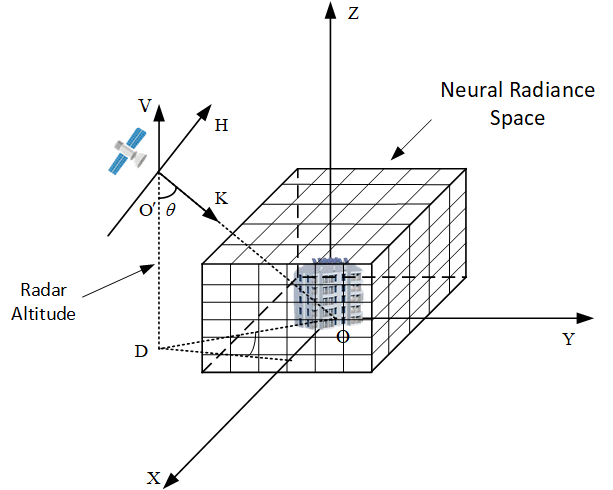}
\caption{Radar observation local coordinate system}
\label{fig_4}
\end{figure}

Consistent with the previous section, let's assume that the number of sampling points in the radar's incident angle, azimuth direction, and slant range direction is denoted as $N_\theta$, $N_a$, and $N_r$, respectively. The corresponding sampling intervals are $\Delta \theta$, $\Delta a$, and $\delta r$. Taking the $i$-th sampling point unit along the $O^\prime H$ direction as the source point of the ray, its coordinates can be written as follows:
\begin{equation}
    {\overrightarrow{v}}_i=\left(-\frac{N_a+1}{2}+i\right)\Delta a\overrightarrow{h}\label{con:12}
\end{equation}

The cell will emit $N_\theta$ rays, and the unit vector ${\overrightarrow{k}}_1$ corresponding to $\theta_1$ is expressed as follows:
\begin{equation}
    {\overrightarrow{k}}_1=\overrightarrow{k}R\left(\theta_1-\theta\right)\label{con:13}
\end{equation}

In Eqs.(\ref{con:13}), $\theta$ represents the radar's pitch angle, and $R(\theta^\prime)$ denotes the affine transformation corresponding to rotating around $OH$ by an angle $\theta^\prime$, given by the equation:
\begin{equation}
    R\left(\theta^\prime\right)=\left[\begin{matrix}1&0&0\\
    0&\cos{\theta^\prime}&\sin{\theta^\prime}\\
    0&-\sin{\theta^\prime}&\cos{\theta^\prime}\\ \end{matrix}\right]\label{con:14}
\end{equation}

The direction of the $\left(i,k\right)$-th ray can be represented as:
\begin{equation}
    {\overrightarrow{d}}_k={\overrightarrow{k}}_1 R \left[ \left(k - \frac{1}{2} \right) \Delta \theta \right]\label{con:15}
\end{equation}

Combining Eqs. (\ref{con:12}) and (\ref{con:15}), we can obtain the representation of the $(i,j,k)$-th sampling point, with its center coordinates given by:
\begin{equation}
    {\overrightarrow{v}}_{i,j,k}={\overrightarrow{v}}_i+{\overrightarrow{d}}_k \left[r + \left(-\frac{N_r+1}{2}+j \right)\Delta r \right]\label{con:16}
\end{equation}

By utilizing affine transformations, we can transform the radar coordinate systems at different observation angles into the same world coordinate system. The transformation equations between the world coordinate system and the radar coordinate system are as follows:
\begin{equation}
    \overrightarrow{v}=R_r \overrightarrow{v}_r + P_r\label{con:17}
\end{equation}
\begin{equation}
    R_r=\left[\begin{matrix}
    -\cos\varphi&-\cos\theta \sin\varphi&-\sin\theta \sin\varphi\\
    0&\sin\theta&-\cos\theta\\
    \sin\varphi&-\cos\theta \cos\varphi&-\sin\theta \cos\varphi\\
    \end{matrix}\right]\label{con:18}
\end{equation}

\noindent Where $\overrightarrow{v}$ and ${\overrightarrow{v}}_r$ represent the coordinates of the same point in the world coordinate system and the radar coordinate system, respectively. $P_r$ represents the coordinates of the origin of the radar coordinate system in the world coordinate system.

Based on this, the correspondence between the radar grid coordinate index and the radar coordinate system is derived. Let's assume that the middle position of the azimuth corresponds to the origin $O$ in the grid coordinate system. Therefore, the spatial position of each grid coordinate $\left(m_i,p_j,q_k\right)$ can be represented as follows:
\begin{equation}
    m_i=-\left(\frac{N_a}{2}\right)+i\label{con:19}
\end{equation}
\begin{equation}
    p_j = \frac{r_{{\rm min}}}{\Delta} + j -1\label{con:20}
\end{equation}
\begin{equation}
    q_k=\frac{N_\theta\theta_1}{\theta_2-\theta_1}+k-1\label{con:21}
\end{equation}

By using Eqs. (\ref{con:19}), (\ref{con:20}), and (\ref{con:21}), we can obtain the coordinates corresponding to the grid index $\left(m_i,p_j,q_k\right)$ as follows:
\begin{equation}
    {\overrightarrow{v}}_{i,j,k}=m_i\Delta x\overrightarrow{h} + \overrightarrow{k}_1 \mathbf{R}\left(k\Delta \theta \right)\left(r_{{\rm min}} + j\Delta r\right)\label{con:22}
\end{equation}

\noindent Where $r_{{\rm min}}$ represents the minimum value in the range direction. By substituting $\Delta x=\Delta a$, $\Delta s=r\Delta \theta$, and equation (22) into equation (6), we can obtain the expression for the scattering energy of a single pixel in the SAR image in the coordinate system.
\begin{equation}
\begin{aligned}
    S_{i,j}&=r\Delta a\Delta r\Delta \theta\sum_{k=0}^{N_\theta - 1}\prod_{j^{'} = 0}^{j} \exp\left[-\Delta r k_{e}^{+}\left(v_{i,j^{'},k} \right) \right]\cdot\\
    &\quad P\left(v_{i,j,k} \right)\prod_{j^{'} = j}^{0} \exp\left[-\Delta r k_{e}^{-}\left(v_{i,j^{'},k} \right) \right]\label{con:23}
\end{aligned}
\end{equation}

\subsection{SAR Neural Radiance Fields}
In the field of optical imaging, the NeRF model introduced an implicit representation approach, which integrates the physical model of volumetric rendering into a neural network. This enables dense reconstruction of 3D voxels from multi-view observed images and further synthesis of new view images using the imaging model. The SAR-NeRF novel proposed in this paper draws inspiration from this idea. It uses a neural network to establish an implicit representation of SAR 3D voxels, encoding the distribution of attenuation coefficient $\sigma_{i,j,k}$ and scattering intensity $S_{i,j,k}$ in the space. The SAR-NeRF are depicted in Fig.5. The attenuation coefficient $\sigma_{i,j,k}$ is related to the voxel's position. If the voxel is located inside the target,$ \sigma_{i,j,k}$ is relatively small, while if the voxel is located outside the target, $\sigma_{i,j,k}$ tends to approach 0. On the other hand, the scattering intensity $S_{i,j,k}$ depends not only on the voxel's position but also on the ray's direction. If the voxel has a smaller angle with the ray, it will have a higher scattering intensity. The representations of $\sigma_{i,j,k}$ and $S_{i,j,k}$ are given as follows:
\begin{equation}
    \sigma_{i,j,k}=F\left(v_{i,j,k}\right)\label{con:24}
\end{equation}
\begin{equation}
    S_{i,j,k}=G\left(v_{i,j,k},d_{i,j,k}\right)\label{con:25}
\end{equation}
\noindent where $v$ and $d$ represent the position and direction of the sampling point and $\gamma \left( p \right)$ is as follows:
\begin{equation}
    \gamma \left( p \right)=\left[ \sin \left( {{2}^{0}}\pi p \right),\cos \left( {{2}^{0}}\pi p \right)\cdots \sin \left( {{2}^{L-1}}\pi p \right),\cos \left( {{2}^{L-1}}\pi p \right) \right]\label{con:new1}
\end{equation}


\begin{figure}[!t]
\centering
\includegraphics[width=3.5in]{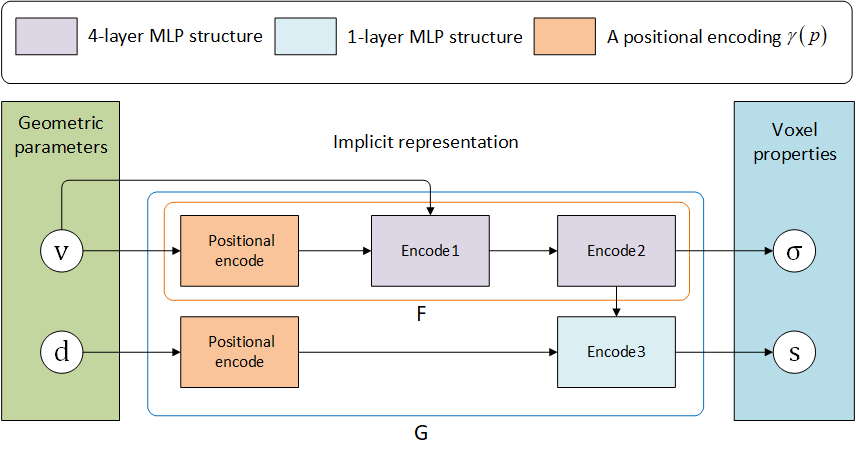}
\caption{SAR Neural Radiance Fields.}
\label{fig_sim}
\end{figure}

Furthermore, this paper has also devised a tailored design for the activation function of the $F\left(v_{i,j,k}\right)$ structure. Let us consider two sampling points placed inside and outside the target, denoted as $v_1$ and $v_2$, respectively. It is evident that $v_1$ is located within the target, implying an inevitable attenuation with $\sigma_1<0$. Similarly, $v_2$ resides outside the target, indicating an absence of attenuation with $\sigma_2=0$. Consequently, it can be easily deduced that $\sigma\le0$. Likewise, concerning the scattering intensity $S$, since $v_1$ exists within the target, it undoubtedly contributes to the scattered energy. Conversely, $v_2$, situated outside the target, inevitably lacks any contribution to the scattered energy. Let us assume that the output of the SAR-NeRF, prior to the activation function, is denoted as $\sigma_r$ and $S_r$. With this in mind, we can provide the following expression.

\begin{equation}
f\left(x\right)=\begin{cases}
 \exp\left(\sum_{i=0}^{j} \sigma_i  \right)S_r & \text{ if } v\in T \\
 0 & \text{ if } v\notin  T
\end{cases} \label{con:26}
\end{equation}
 
 Here, $v$ represents the coordinates of a sampling point in space, and $T$ represents the region occupied by the target. Based on Eqs. (\ref{con:26}), we can derive the final output of the SAR-NeRF as follows:
 \begin{equation}
     \sigma=-{\rm ReLU}\left(\sigma_r\right)\label{con:27}
 \end{equation}
 \begin{equation}
     \hat{I}=-\tanh{\left(\sigma_r\right)}\exp{\left(\sum_{i=0}^{j}\sigma_i\right)S_r}\label{con:28}
 \end{equation}

 Since SAR images are functions of voxel density and scattering strength, we can fit the MLP by minimizing the error between the predicted image ${\hat{I}}_{i,j}(\sigma,S)$ and the ground truth image $I_{i,j}$, which can be expressed as:
 \begin{equation}
     \mathop{\arg\min}\limits_{\sigma,S}\frac{1}{N_a N_r}\sum_{i = 1}^{N_a} \sum_{j = 1}^{N_r}\left \| {\hat{I}}_{i,j}\left(\sigma,S\right)-I_{i,j} \right \|_{2}^{2} \label{con:29}
 \end{equation}
 By optimizing the activation function of the network, the convergence speed of the network can be significantly increased. This optimization also addresses the challenge of training SAR-NeRF when the background scattering energy is zero. Moreover, this activation function enhances the differentiation between attenuation coefficients inside and outside the target, facilitating the extraction of geometric models.

 \subsection{Reconstruction of 3D geometric model}
SAR-NeRF accomplishes the prediction of attenuation coefficient and scattering intensity distributions in the sampling space. However, there exists a strong correlation between the attenuation coefficient and the voxel distribution of the target in space. As described in the previous section, if a voxel is located inside the target, its attenuation coefficient $\sigma_{i,j,k}$ is greater than zero, while if it is outside the target, $\sigma_{i,j,k}$ equals zero. By leveraging this concept, we can reconstruct the three-dimensional geometric model of the target using SAR-NeRF. Firstly, based on prior knowledge, the neural radiance range of SAR-NeRF, referred to as the neural radiance space, is determined. The voxel distribution in this space is obtained by averaging the samples taken within the neural radiance space. Finally, the voxel information is input into the SAR-NeRF to obtain the distribution of attenuation coefficients in this space. Points with $\sigma_{i,j,k}$ equal to zero are removed, resulting in the three-dimensional voxel model of the target. The specific process is illustrated in Fig.6.
 \begin{figure}[!t]
\centering
\includegraphics[width=3.5in]{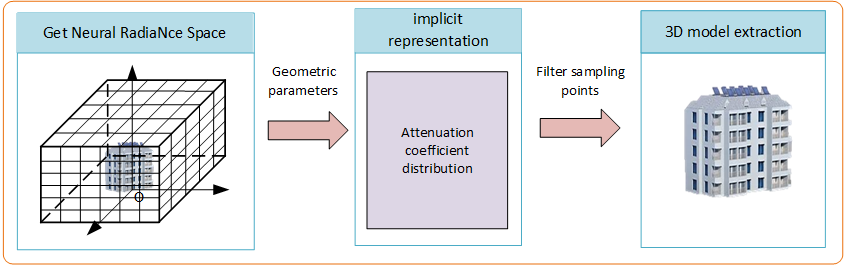}
\caption{Temporary placeholder, later change}
\label{fig_6}
\end{figure}

 \section{Experiments}
The experiments in this study are divided into three main parts: forward rendering experiments, multi-view image generation from rendered images, and multi-view image generation from real SAR images. In the forward rendering tests, the effectiveness of the SAR voxel rendering method is validated. The experiments on multi-view image generation from rendered images demonstrate the reliability of SAR Neural Radiative Field (SAR-NeRF) in the task of generating multi-view images, as well as its ability to learn geometric information of the target. The experiments on multi-view image generation from real SAR images primarily verify the effectiveness of SAR-NeRF on real data.

\subsection{Validation with forward rendering experiment}
We first validated the effectiveness of SAR image voxel rendering. A typical building was chosen as our validation target model, which is simplified as a cuboid structure. Through the mapping and projection mechanism of SAR imaging, we know that the representation of the building in SAR images can be composed of ground scattering (SG), wall scattering (SW), roof scattering (SR), and shadows (S).


\begin{figure}[!t]
\centering
\subfloat[model a and Corresponding voxel rendered image.]{
    \begin{minipage}[t]{0.45\linewidth}
        \centering
        \includegraphics[width=1\linewidth]{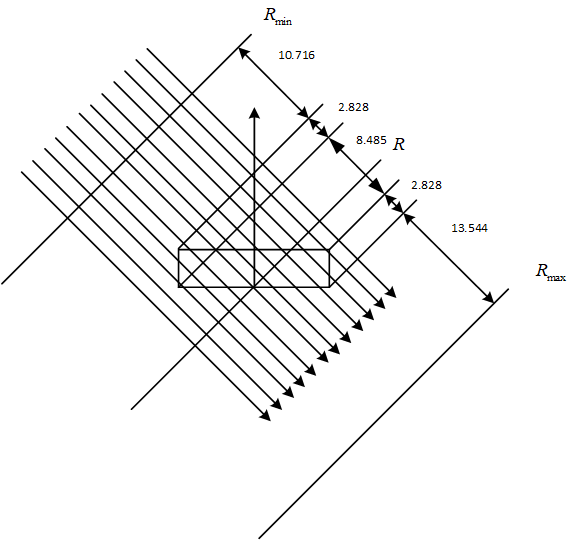}
    \end{minipage}
    \begin{minipage}[t]{0.45\linewidth}
        \centering
        \includegraphics[width=1\linewidth]{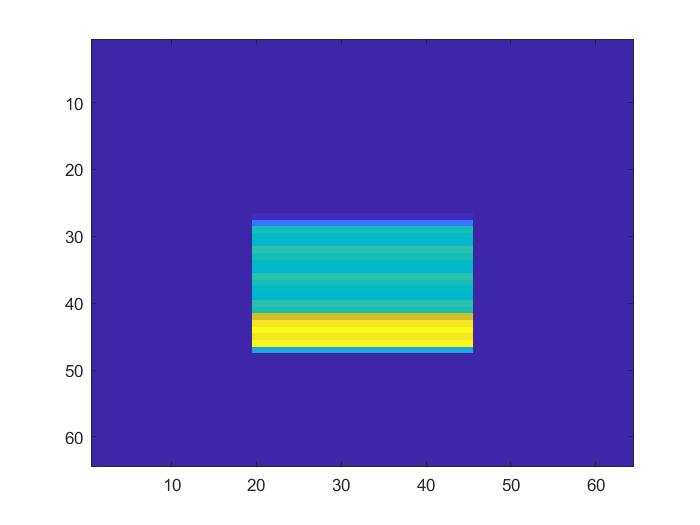}
    \end{minipage}
}\\
\subfloat[model b and Corresponding voxel rendered image.]{
    \begin{minipage}[t]{0.45\linewidth}
        \centering
        \includegraphics[width=1\linewidth]{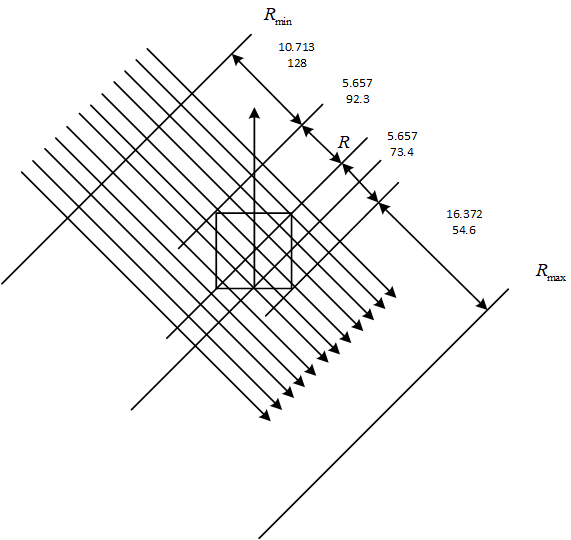}
    \end{minipage}
    \begin{minipage}[t]{0.45\linewidth}
        \centering
        \includegraphics[width=1\linewidth]{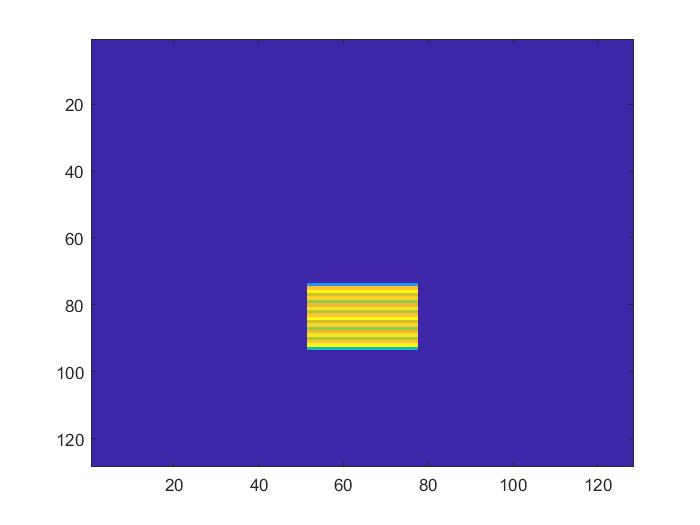}
    \end{minipage}
}\\
\subfloat[model c and Corresponding voxel rendered image.]{
    \begin{minipage}[t]{0.45\linewidth}
        \centering
        \includegraphics[width=1\linewidth]{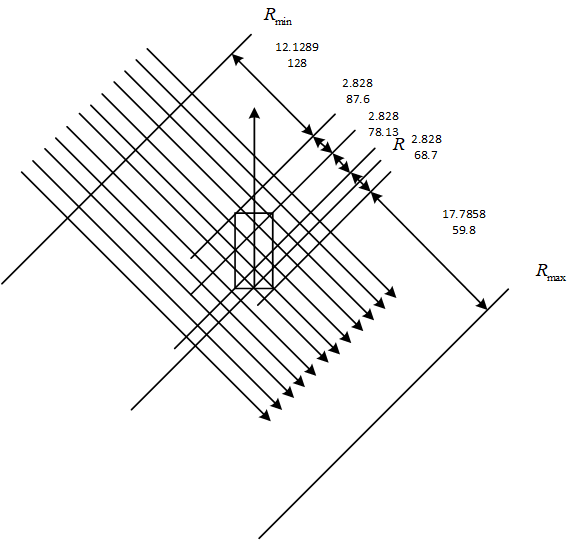}
    \end{minipage}
    \begin{minipage}[t]{0.45\linewidth}
        \centering
        \includegraphics[width=1\linewidth]{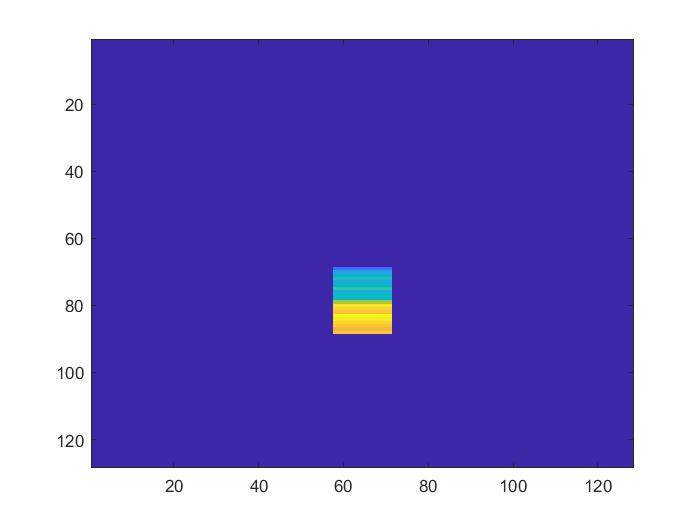}
    \end{minipage}
}
\caption{Under the condition that the pitch angle is 45 degrees and the radar altitude is 10000 meters, the three types of target SAR images rendered by voxel.}
\label{fig.8}
\end{figure}

The distribution of scattering components varies with the changes in wall height ($h$) and roof width ($w$). The model shown in Fig. 7(a) satisfies this condition.
\begin{equation}
    \frac{w}{h}>\cot{\alpha}\label{con:30}
\end{equation}

At this point, we increase the height of the building, causing SW and SR to completely overlap. The model shown in Figure 7(c) satisfies this condition.
\begin{equation}
    \frac{w}{h}=\cot{\alpha}\label{con:31}
\end{equation}

Upon further increasing the height of the building based on Fig. 7(c), the distribution of SW exceeds that of SR. The model depicted in Fig. 7(e) satisfies this condition.
\begin{equation}
    \frac{w}{h}<\cot{\alpha}\label{con:32}
\end{equation}

This experiment provides preliminary evidence that supports the rationality and effectiveness of SAR voxel rendering.

\subsection{Multi-view image generation based on rendered image}
\subsubsection{Dataset}
This section demonstrates the experimental results of multi-view image generation on simulated rendered images. To obtain realistic scattering textures of the models and eliminate background scattering effects, the simulated rendered image dataset includes rendering images of three types of target models: cuboid, upright four-sided platform (with a smaller top and a larger base), and four-sided platform. The images have a size of 128×128 and a spatial resolution of 0.3m/pixel. As shown in Figure 8, the pitch angle of the three instances' rendered images is set to $45^{\circ } $, with a radar altitude of 10,000 meters.

\begin{figure}[!t]
\centering
\subfloat[Voxel rendered image of a cuboid.]{
    \begin{minipage}[t]{0.45\linewidth}
        \centering
        \includegraphics[width=1\linewidth]{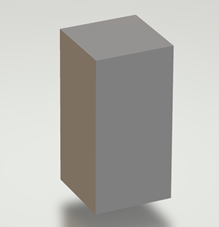}
    \end{minipage}
    \begin{minipage}[t]{0.45\linewidth}
        \centering
        \includegraphics[width=1\linewidth]{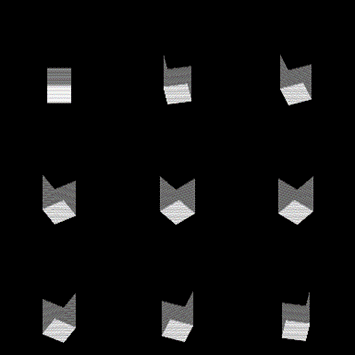}
    \end{minipage}
}\\
\subfloat[Voxel rendering image of the upright four-sided platform model.]{
    \begin{minipage}[t]{0.45\linewidth}
        \centering
        \includegraphics[width=1\linewidth]{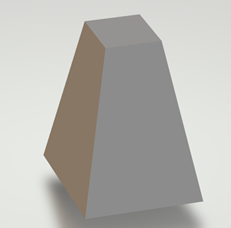}
    \end{minipage}
    \begin{minipage}[t]{0.45\linewidth}
        \centering
        \includegraphics[width=1\linewidth]{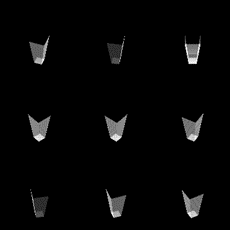}
    \end{minipage}
}\\
\subfloat[Voxel rendered image of the inverted four-sided platform model.]{
    \begin{minipage}[t]{0.45\linewidth}
        \centering
        \includegraphics[width=1\linewidth]{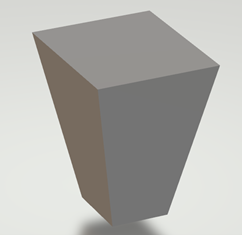}
    \end{minipage}
    \begin{minipage}[t]{0.45\linewidth}
        \centering
        \includegraphics[width=1\linewidth]{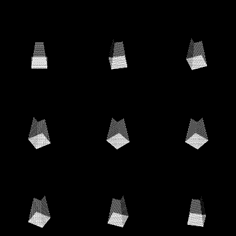}
    \end{minipage}
}
\caption{Under the condition that the pitch angle is 45 degrees and the radar altitude is 10000 meters, the three types of target SAR images are rendered by voxel.}
\label{fig.8}
\end{figure}

\subsubsection{Evaluation metrics}
For evaluating the performance of multi-view image generation using SAR-NeRF, besides visual comparisons between the generated images and the ground truth datasets, we utilize the peak signal-to-noise ratio (PSNR) and the Learned Perceptual Image Patch Similarity (LPIPS) as quantitative evaluation metrics. PSNR is a distortion-based metric that tends to favor smooth or blurry reconstruction results. As a complement, LPIPS calculates the perceptual distance between two images, which is more aligned with human perception. In our case, we use AlexNet to extract features for LPIPS calculation.
\subsubsection{Experiment set up}
Subsequently, we conducted tests on the simulated rendered SAR images of the three model categories. For each model category, under the observation conditions of a radar height of 10,000 meters (default setting in this experiment) and a pitch angle of $45^{\circ}$, we selected 36 images from the azimuth angle range of $0^{\circ}$,$359^{\circ}$ with a $10^{\circ}$ interval as the training set, while the remaining 324 images were used as the test set. Firstly, we examined the similarity of the generated samples when applied to the test set. Figure 9 showcases the visual comparison between the SAR-NeRF generated synthesized perspective images (lower row of each panel) and the corresponding original rendered ground truth images from the test set (upper row of each panel). In order to assess the SAR-NeRF's generalization ability to arbitrary angles, the leftmost and rightmost images in each panel, marked with red boxes, represent the SAR rendered images at azimuth angles of $0^{\circ}$ and $90^{\circ}$ (which belong to the training set), respectively, while the intermediate angles have an interval of $10^{\circ}$. It can be observed that as the radar observation angle varies, SAR-NeRF is capable of inferring the corresponding changes in the synthesized rendered SAR images and reasonably outputs the distribution of scattered energy in the synthesized images.
\begin{figure*}[!t]
    \centering
    \subfloat[Comparison chart of cuboid]{
        \begin{minipage}[t]{0.8\linewidth}
            \centering
            \includegraphics[width=1\linewidth]{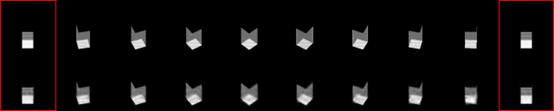}
        \end{minipage}
    }\\
    \subfloat[Comparison chart of upright four-sided platform]{
        \begin{minipage}[t]{0.8\linewidth}
            \centering
            \includegraphics[width=1\linewidth]{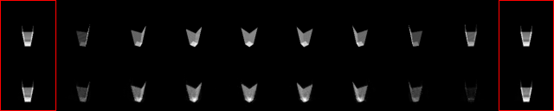}
        \end{minipage}
    }\\
    \subfloat[Comparison chart of inverted four-sided platform]{
        \begin{minipage}[t]{0.8\linewidth}
            \centering
            \includegraphics[width=1\linewidth]{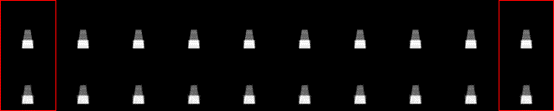}
        \end{minipage}
    }\\
    \caption{The comparison between the SAR image generated by the SAR-NeRF and the SAR image obtained by voxel rendering, the previous row is the true value, and the next row is the generated SAR image.}
    \label{Fig.9}
\end{figure*}

To quantitatively evaluate the generation performance, we varied the angle interval/training set size in the aforementioned experiments. The quantitative comparisons are presented in Table 1. Overall, our model achieves excellent performance with a reasonable amount of data, and even with a reduced data size, SAR-NeRF still demonstrates commendable performance.

\begin{table}[!t]
\caption{The influence of the number of training samples on the evaluation index under the condition of the fixed azimuth angle in SAR-NERF\label{tab:table1}}
\centering
\begin{tabular}{|c|c|cc|}
\hline
\multirow{2}{*}{Model type}                                                              & \multirow{2}{*}{Angle interval datavolume} & \multicolumn{2}{c|}{Evaluation Metrics} \\ \cline{3-4} 
                                                                                         &                                            & \multicolumn{1}{c|}{PSNR}     & LPIPS   \\ \hline
\multirow{4}{*}{cuboid}                                                                  & 5/72                                       & \multicolumn{1}{c|}{34.8121}  & 0.0153  \\ \cline{2-4} 
                                                                                         & 10/36                                      & \multicolumn{1}{c|}{32.8659}  & 0.0358  \\ \cline{2-4} 
                                                                                         & 15/24                                      & \multicolumn{1}{c|}{31.6574}  & 0.0549  \\ \cline{2-4} 
                                                                                         & 30/12                                      & \multicolumn{1}{c|}{27.3821}  & 0.0711  \\ \hline
\multirow{4}{*}{\begin{tabular}[c]{@{}c@{}}upright four-\\ sided platform\end{tabular}}  & 5/72                                       & \multicolumn{1}{c|}{34.8916}  & 0.0305  \\ \cline{2-4} 
                                                                                         & 10/36                                      & \multicolumn{1}{c|}{34.6709}  & 0.0334  \\ \cline{2-4} 
                                                                                         & 15/24                                      & \multicolumn{1}{c|}{32.8216}  & 0.0404  \\ \cline{2-4} 
                                                                                         & 30/12                                      & \multicolumn{1}{c|}{26.3013}  & 0.0962  \\ \hline
\multirow{4}{*}{\begin{tabular}[c]{@{}c@{}}inverted four-\\ sided platform\end{tabular}} & 5/72                                       & \multicolumn{1}{c|}{35.7120}  & 0.0198  \\ \cline{2-4} 
                                                                                         & 5/72                                       & \multicolumn{1}{c|}{34.7120}  & 0.0297  \\ \cline{2-4} 
                                                                                         & 15/24                                      & \multicolumn{1}{c|}{31.1112}  & 0.0506  \\ \cline{2-4} 
                                                                                         & 30/12                                      & \multicolumn{1}{c|}{26.6266}  & 0.0877  \\ \hline
\end{tabular}
\end{table}

Based on the aforementioned experiments, this study further validates the learning capability of SAR-NeRF for voxel distribution in the sampling space by simultaneously varying the azimuth and pitch angles. We extended the pitch angle range from the original $45^{\circ}$ to $\left[35^{\circ}, 55^{\circ}\right]$. For the training set, we selected all odd pitch angles in the range $\left[35^{\circ}, 55^{\circ}\right]$ and varied the azimuth angle from $0^{\circ}$ to $360^{\circ}$ with a $10^{\circ}$ interval. For the interpolation experiment, we used all even pitch angles in the range $\left[35^{\circ}, 55^{\circ}\right]$ and varied the azimuth angle, which served as the unseen test set. Additionally, we generated images with pitch angles of $30^{\circ}$ and $60^{\circ}$ as the extrapolation experiment's unseen test set. Figure 10 and Figure 11 respectively demonstrate the results of the interpolation and extrapolation experiments on the cuboid model.
\begin{figure*}[!t]
    \centering
    \subfloat[SAR-NeRF Generated Images]{
        \begin{minipage}[t]{0.8\linewidth}
            \centering
            \includegraphics[width=1\linewidth]{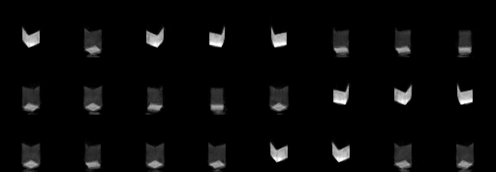}
        \end{minipage}
    }\\
    \subfloat[SAR voxel rendered image]{
        \begin{minipage}[t]{0.8\linewidth}
            \centering
            \includegraphics[width=1\linewidth]{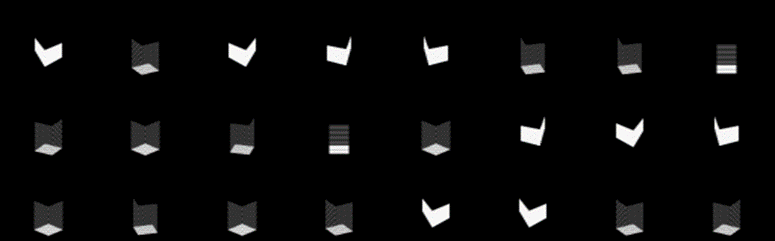}
        \end{minipage}
    }\\
    \caption{Test results of interpolation experiment}
    \label{Fig.10}
\end{figure*}

\begin{figure}[!t]
    \centering
    \subfloat[The image generated by SAR-NeRF at an pitch angle of $60^{\circ}$.]{
        \begin{minipage}[t]{0.8\linewidth}
            \centering
            \includegraphics[width=1\linewidth]{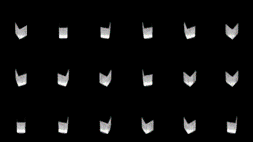}
        \end{minipage}
    }\\
    \subfloat[Voxel rendering image at a pitch angle of $60^{\circ}$.]{
        \begin{minipage}[t]{0.8\linewidth}
            \centering
            \includegraphics[width=1\linewidth]{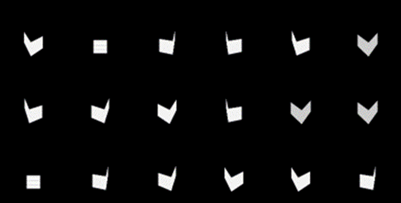}
        \end{minipage}
    }\\
    \subfloat[The image generated by SAR-NeRF at an pitch angle of $30^{\circ}$.]{
        \begin{minipage}[t]{0.8\linewidth}
            \centering
            \includegraphics[width=1\linewidth]{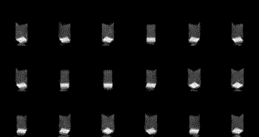}
        \end{minipage}
    }\\
    \subfloat[Voxel rendering image at a pitch angle of $30^{\circ}$.]{
        \begin{minipage}[t]{0.8\linewidth}
            \centering
            \includegraphics[width=1\linewidth]{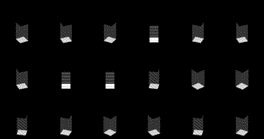}
        \end{minipage}
    }\\
    
    \caption{Test results of extrapolation experiments.}
    \label{Fig.11}
\end{figure}

Building upon the interpolation and extrapolation experiments with the pitch angle, this study proceeded with the extraction of three-dimensional voxels. The selected sampling space had dimensions of 20×20×20m³, and it was uniformly sampled to obtain 256×256×256 sample points. These sample points were then fed into the SAR-NeRF model trained in the previous step for prediction. Figure 12 illustrates the results of the voxel extraction process, showcasing the effectiveness of the approach in generating three-dimensional models.
 \begin{figure}[!t]
\centering
\includegraphics[width=3in]{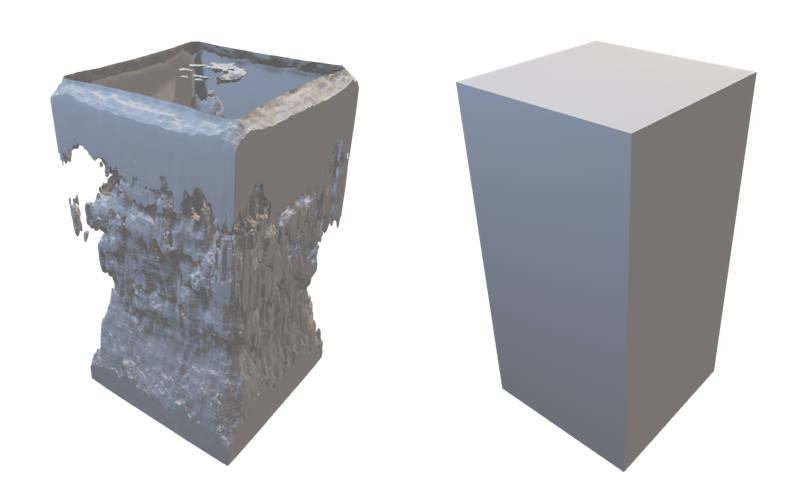}
\caption{3D model extraction experiment results.}
\label{fig_12}
\end{figure}

Building upon the previous experiments, this study further conducted experiments on the upright pyramid and inverted pyramid models. Figure 13 presents the three-dimensional voxel extraction results for the pyramid models, demonstrating the effectiveness of the approach.
\begin{figure}[!t]
\centering
\subfloat[3D model extraction results.]{
    \begin{minipage}[t]{0.8\linewidth}
        \centering
        \includegraphics[width=1\linewidth]{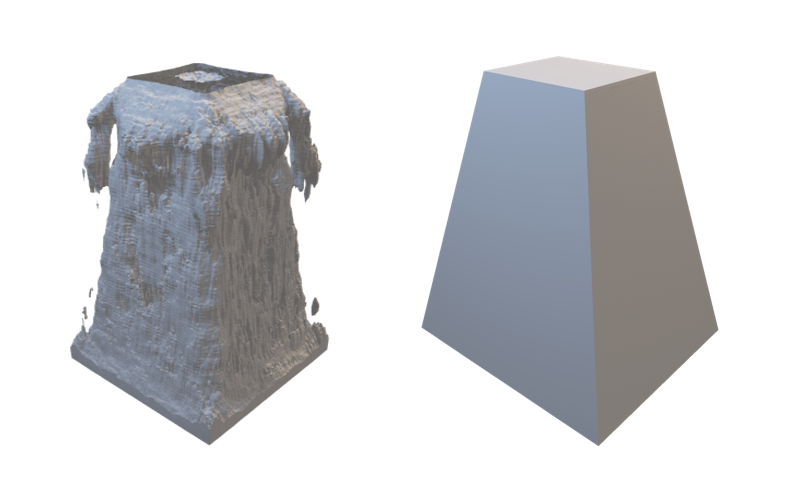}
    \end{minipage}
}\\
\subfloat[3D model extraction results.]{
    \begin{minipage}[t]{0.8\linewidth}
        \centering
        \includegraphics[width=1\linewidth]{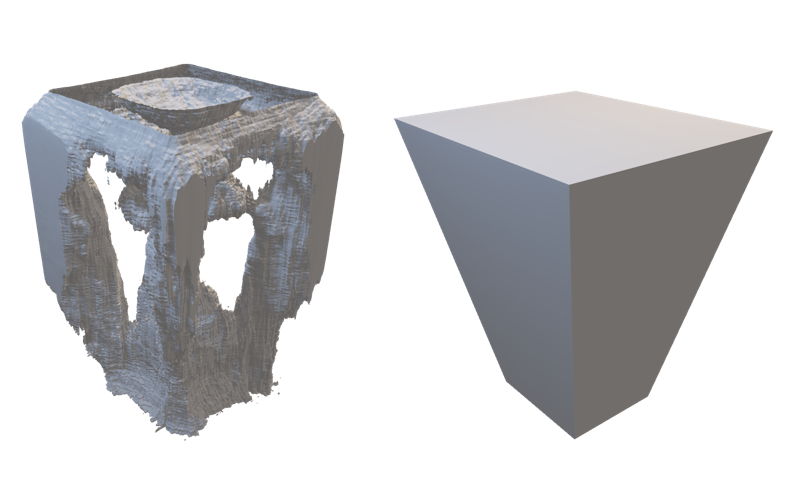}
    \end{minipage}
}
\caption{More 3D model extraction results.}
\label{fig.13}
\end{figure}

\subsection{Multi-view image generation on MSTAR dataset}
\subsubsection{dataset}
In this section, the effectiveness of SAR Neural Radiance Fields (SAR-NeRF) on real SAR image datasets was further validated. The dataset used in this section is the Moving and Stationary Target Acquisition and Recognition (MSTAR) dataset, which provides high-resolution real SAR data. The 2S1 vehicle in the dataset was selected as the experimental target. The SAR-NeRF model was evaluated for multi-view image generation from a single pitch angle and multiple azimuth angles, as well as from multiple pitch angles and multiple azimuth angles. It is important to note that the original amplitude values of the MSTAR data were used in this study, and the image size was set to 128×128. Therefore, the generated images by SAR-NeRF retained the statistical characteristics of the real SAR images.

\subsubsection{technical details}
In the scenario of a single pitch angle, we have selected the MSTAR dataset with an pitch angle of 17 degrees. From this dataset, we have chosen 299 images as our samples. Specifically, we have taken 24 images with azimuth angles spaced 15 degrees apart as the training set, while the remaining 275 images form the testing set. Figure 14 showcases the results of SAR Neural Radiance Field (NeRF) generation on the MSTAR dataset.
\begin{figure*}[!t]
    \centering
    \subfloat[SAR-NeRF Generated Images]{
        \begin{minipage}[t]{0.8\linewidth}
            \centering
            \includegraphics[width=1\linewidth]{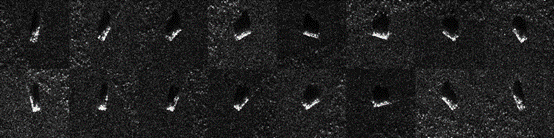}
        \end{minipage}
    }\\
    \subfloat[SAR voxel rendered image]{
        \begin{minipage}[t]{0.8\linewidth}
            \centering
            \includegraphics[width=1\linewidth]{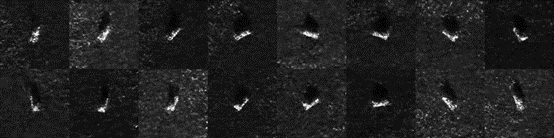}
        \end{minipage}
    }\\
    \caption{Generation Effect of SAR-NeRF on MSTAR Dataset}
    \label{Fig.14}
\end{figure*}

The observation from Figure 14 reveals that the SAR Neural Radiance Field (NeRF) not only accomplishes a fine fit to the target details but also reconstructs the shadow regions of the objects. This further validates the effectiveness and practicality of our SAR-NeRF voxel rendering approach based on mapping projection principles. Building upon this experiment, we further explored azimuth angle intervals of $5^{\circ}$, $10^{\circ}$, and $15^{\circ}$. The experimental results substantiate that SAR-NeRF maintains impressive performance even when the sample quantity undergoes variations, as illustrated in Table 2.  
 \begin{table}[!t]
 \caption{Peak signal-to-noise ratio of generated images and real images\label{tab:table2}}
\begin{tabular}{|c|c|c|c|}
\hline
datasets & Angular interval 5 & Angular interval 10 & Angular interval 15 \\ \hline
2S1      & 31.26              & 30.31               & 30.15               \\ \hline
ZSU234   & 30.9778            & 30.5655             & 30.00               \\ \hline
BRDM2    & 32.74              & 32.14               & 30.72               \\ \hline
A64      & 30.5255            & 29.9982             & 29.95               \\ \hline
\end{tabular}
\end{table}

Building upon the previous experiments, this study further investigates the multi-view image generation capabilities of the SAR Neural Radiance Field (NeRF) under varying pitch angles. Specifically, we selected the 2S1 vehicle and utilized data at pitch angles of $15^{\circ}$, $17^{\circ}$, and $30^{\circ}$, with azimuth angle intervals of $5^{\circ}$, as the training set for SAR-NeRF. The remaining data, which were not used during training, were employed for azimuth angle interpolation experiments. The results of these experiments are depicted in Figure 15. Additionally, a set of data was generated at pitch angles of $12^{\circ}$,$20^{\circ}$,$35^{\circ}$, where corresponding real SAR images were not available. Therefore, only the generated results are showcased in Figure 16.

\begin{figure}[!t]
\centering
\subfloat[ground truth.]{
    \begin{minipage}[t]{0.8\linewidth}
        \centering
        \includegraphics[width=1\linewidth]{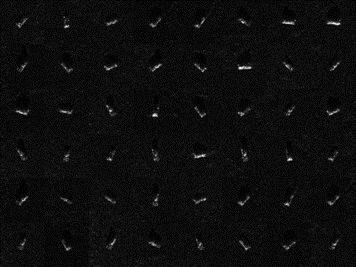}
    \end{minipage}
}\\
\subfloat[Output of SAR-NeRF]{
    \begin{minipage}[t]{0.8\linewidth}
        \centering
        \includegraphics[width=1\linewidth]{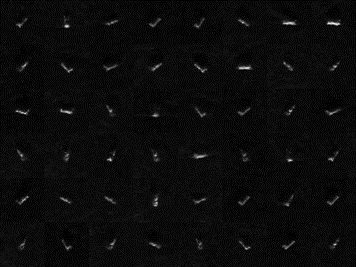}
    \end{minipage}
}
\caption{Test results under multiple pitch angle data sets}
\label{fig.15}
\end{figure}

\begin{figure}[!t]
\centering
\subfloat[The generated image at a pitch angle of $12^{\circ}$]{
    \begin{minipage}[t]{0.9\linewidth}
        \centering
        \includegraphics[width=1\linewidth]{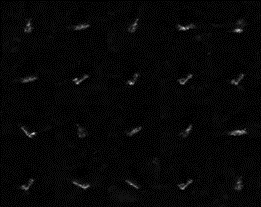}
    \end{minipage}
}\\
\subfloat[The generated image at a pitch angle of $20^{\circ}$]{
    \begin{minipage}[t]{0.9\linewidth}
        \centering
        \includegraphics[width=1\linewidth]{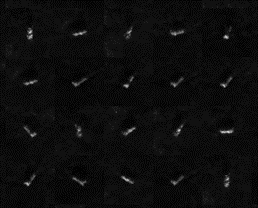}
    \end{minipage}
}\\
\subfloat[The generated image at a pitch angle of $35^{\circ}$]{
    \begin{minipage}[t]{0.9\linewidth}
        \centering
        \includegraphics[width=1\linewidth]{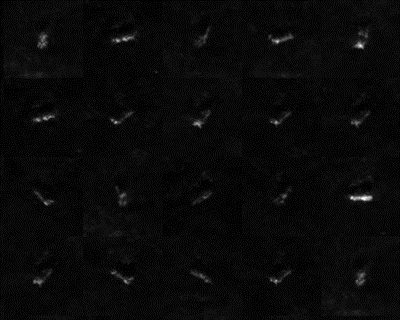}
    \end{minipage}
}
\caption{Interpolation and extrapolation test results of pitch angle}
\label{fig.16}
\end{figure}

The aforementioned experiments provide evidence that the SAR Neural Radiance Field (NeRF) is capable of learning the spatial geometric information of the target scene, rather than simply interpolating between adjacent angles for smooth effects. In addition, this study also explored the extraction of three-dimensional models from the MSTAR dataset. However, progress in this aspect was limited due to the challenges posed by background noise, clutter, and variations in the background within the dataset, as well as the limited coverage of pitch angles.

\subsection{Data augumentation using SAR-NeRF under FSL conditions}
To validate the performance improvement of using NeRF-generated data for FSL classification tasks, we conducted experiments using ten classes of MSTAR data as the dataset for the classification network, with ResNet50 as the classification network. First, we selected {36, 24, 12, 8, 4} SAR images from each class of MSTAR data to create five sets of test samples. We trained the classification network on these samples to obtain the classification accuracy without using SAR-NeRF-enhanced data. Next, we used these five sets of test samples as training data for SAR-NeRF and generated MSTAR data from additional angles. We then used this augmented data as the training set for the classification network and obtained the classification accuracy with SAR-NeRF-enhanced data. Table 4 presents the impact of using SAR-NeRF-enhanced data on the classification accuracy under the FSL conditions.
\begin{table}[!t]
\centering
\caption{Effect of SAR-NeRF enhanced data on network classification accuracy\label{tab:table3}}
\begin{tabular}{|c|cc|}
\hline
shot    & \multicolumn{2}{c|}{Accuracy of Test Samples}            \\ \hline
        & \multicolumn{1}{c|}{without SAR-NeRF} & with SAR-NeRF \\ \hline
36-shot & \multicolumn{1}{c|}{97.6\%}           & 97.8\%           \\ \hline
24-shot & \multicolumn{1}{c|}{95.1\%}           & 95.6\%           \\ \hline
12-shot & \multicolumn{1}{c|}{85.2\%}           & 91.6\%           \\ \hline
8-shot  & \multicolumn{1}{c|}{72.9\%}           & 78.9\%           \\ \hline
4-shot  & \multicolumn{1}{c|}{60.1\%}           & 65.2\%           \\ \hline
\end{tabular}
\end{table}

From this table, it can be observed that when the training sample size is 36 shots and 24 shots, the classification accuracy without using augmented data already exceeds 95\%, showing only a small difference compared to the classification accuracy with SAR-NeRF-enhanced data. However, as the training sample size decreases to 12 shots and 8 shots, the classification accuracy with SAR-NeRF-enhanced data exhibits a significant improvement over the accuracy without augmented data. When the training sample size reduces to 4 shots, although the classification accuracy with SAR-NeRF-enhanced data still shows improvement, the magnitude of improvement decreases. In summary, SAR-NeRF can significantly enhance the performance of the classification task when the training sample size is appropriate. However, when the training sample size is already sufficient, the original classification network is capable of meeting the classification performance requirements. On the other hand, when the training sample size is extremely limited, the generation capability of SAR-NeRF itself may be compromised, resulting in a diminished impact on the performance enhancement of the classification network.

\bibliographystyle{IEEEtran}
\bibliography{egbib}

\end{document}